\documentclass[lettersize,journal]{IEEEtran}
\usepackage{amsmath}
\usepackage{amsfonts} 
\usepackage{algorithmic}
\usepackage{array}
\usepackage[caption=false,font=normalsize,labelfont=sf,textfont=sf]{subfig}
\usepackage{textcomp}
\usepackage{stfloats}
\usepackage{url}
\usepackage{verbatim}
\usepackage{graphicx}
\usepackage{bm}
\usepackage{graphicx}
\usepackage{booktabs}
\usepackage{subcaption}
\usepackage{colortbl}
\usepackage{threeparttable}
\usepackage{makecell}
\usepackage{cancel}
\usepackage{bbding}
\usepackage{color}
\usepackage{multirow}
\usepackage{hyperref}
\usepackage{orcidlink}

\usepackage{hyperref}
\hypersetup{
colorlinks=true,
urlcolor=magenta,
pdftitle={Overleaf Example},
pdfpagemode=FullScreen,
}

\usepackage[hang,flushmargin]{footmisc}
\usepackage{cases}
\PassOptionsToPackage{table}{xcolor}
\usepackage[table]{xcolor}
\definecolor{lightgray}{RGB}{220, 220, 220}
\definecolor{lightgreen}{RGB}{216, 237, 223}
\graphicspath{{Figure/}}  
\newcommand{\etal}{\textit{et al}.}
\newcommand{\ie}{\textit{i}.\textit{e}.}
\newcommand{\eg}{\textit{e}.\textit{g}.}
\newcommand{\vs}{\textit{v}\textit{s}.}

\newcommand{\model}{{T2S-QA}}
\newcommand{\dataset}{{ViTXT-GQA}}

\def\BibTeX{{\rm B\kern-.05em{\sc i\kern-.025em b}\kern-.08em
    T\kern-.1667em\lower.7ex\hbox{E}\kern-.125emX}}
\usepackage{balance}

\begin{document}
\title{Scene-Text Grounding for Text-Based Video Question Answering}
\author{Sheng Zhou$^{\orcidlink{0009-0007-4215-5464}}$, Junbin Xiao$^{\orcidlink{0000-0001-5573-6195}}$, Xun Yang$^{\orcidlink{0000-0003-0201-1638}}$, Peipei Song$^{\orcidlink{0000-0001-6764-3375}}$, Dan Guo$^{\orcidlink{0000-0003-2594-254X}}$, \emph{Member, IEEE}, \\
Angela Yao$^{\orcidlink{0000-0001-7418-6141}}$, \emph{Member, IEEE}, Meng Wang$^{\orcidlink{0000-0002-3094-7735}}$, \emph{IEEE Fellow}, Tat-Seng 
Chua$^{\orcidlink{0000-0001-6097-7807}}$

\thanks{This work was supported in part by the National Natural Science Foundation of China (U22A2094, 62272144, 72188101, 62020106007, and 62272435), the Major Project of Anhui Province (2408085J040, 202203a05020011), and the Fundamental Research Funds for the Central Universities (JZ2024HGTG0309, JZ2024AHST0337, and JZ2023YQTD0072).
\emph{(Corresponding authors: Junbin Xiao; Xun Yang; Dan Guo.
)}

Sheng Zhou, Dan Guo, and Meng Wang are with the School of Computer Science and Information Engineering, Hefei University of Technology, Hefei, 230601, China (email: hzgn97@gmail.com; guodan@hfut.edu.cn; eric.mengwang@gmail.com). \\ 
Junbin Xiao, Angela Yao, and Tat-Seng 
Chua are with the School of Computing, National University of Singapore, 117418, Singapore. (email: \{junbin, ayao, chuats\}@comp.nus.edu.sg). \\
Xun Yang and Peipei Song are with the School of Information Science and Technology, University of Science and Technology of China, Hefei, 230026, China. (email: xyang21@ustc.edu.cn; beta.songpp@gmail.com). \\
}}

\markboth{IEEE TRANSACTIONS ON MULTIMEDIA,~Vol.~XX, No.~X, 2025}%
{How to Use the IEEEtran \LaTeX \ Templates}

\maketitle

\begin{abstract}
\label{sec:abs}
Existing efforts in text-based video question answering (TextVideoQA) are criticized for their opaque decision-making and heavy reliance on scene-text recognition. In this paper, we propose to study \emph{Grounded TextVideoQA} by forcing models to answer questions and spatio-temporally localize the relevant scene-text regions, thus decoupling QA from scene-text recognition and promoting research towards interpretable QA. The task has three-fold significance. 
First, it encourages scene-text evidence versus other short-cuts for answer predictions. Second, it directly accepts scene-text regions as visual answers, thus circumventing the problem of ineffective answer evaluation by stringent string matching. Third, it isolates the challenges inherited in VideoQA and scene-text recognition. This enables the diagnosis of the root causes for failure predictions, \eg, wrong QA or wrong scene-text recognition?
To achieve Grounded TextVideoQA, we propose the \model~model that highlights a disentangled temporal-to-spatial contrastive learning strategy for weakly-supervised scene-text grounding and grounded TextVideoQA. To facilitate evaluation,  we construct a new dataset \emph{\dataset}~which features 52\textbf{K} scene-text bounding boxes within 2.2\textbf{K} temporal segments related to 2\textbf{K} questions and  729 videos. With \dataset, we perform extensive experiments and demonstrate the severe limitations of existing techniques in Grounded TextVideoQA. While \model~achieves superior results, 
the large performance gap with human leaves ample space for improvement. Our further analysis of oracle scene-text inputs posits that the major challenge is scene-text recognition.
To advance the research of Grounded TextVideoQA, our dataset and code are at \url{https://github.com/zhousheng97/ViTXT-GQA.git}

\end{abstract}


%------------------------------------------------------
\begin{figure}
\centering
\includegraphics[width=\columnwidth]{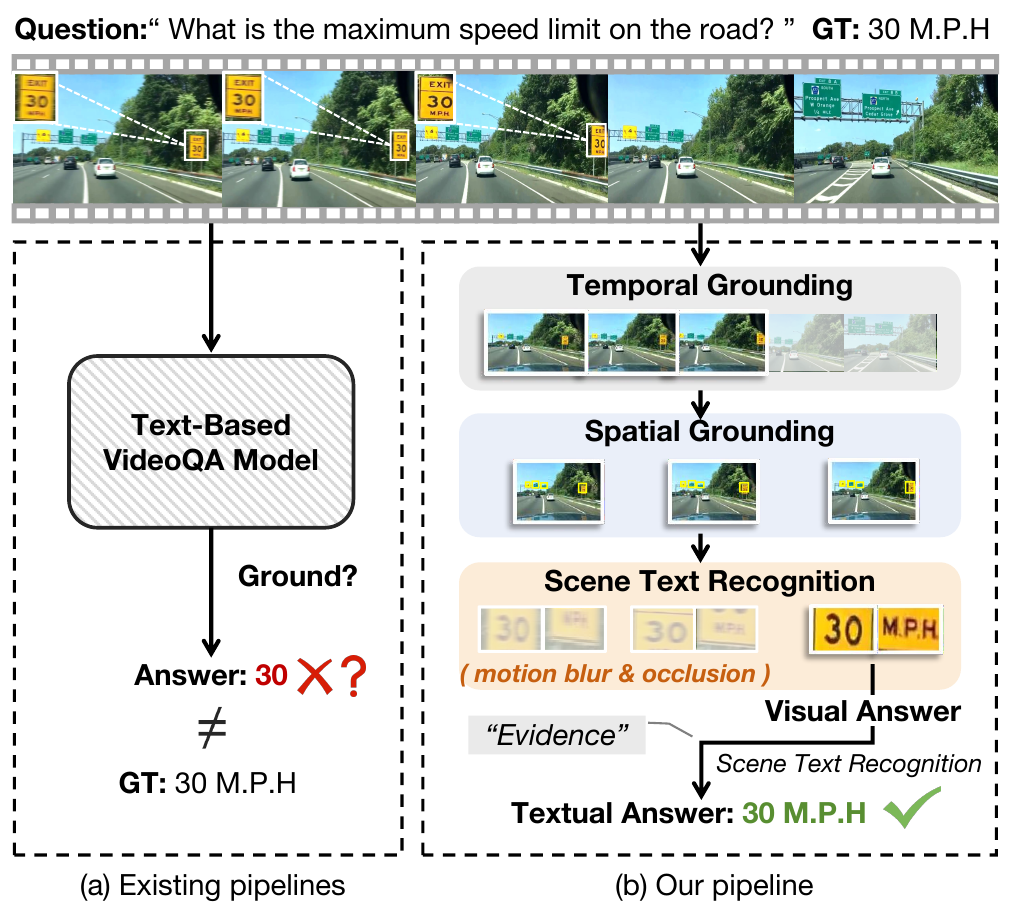}
% \vspace{-0.2in}
\caption{
{Comparison between existing research and our work for TextVideoQA.} (a) Existing research has two major problems: 1) Opaque decision-making; they hardly tell if their answers (${\textit{e.g.}}$, “30”) are originated from the relevant scene texts in the videos,  or attributed to other short-cuts. 2) Heavy reliance on scene-text recognition; their low QA accuracy could be due to a failure in decoding the textual answer (${\textit{e.g.}}$, ``30 M.P.H.'') from the corresponding scene text region. (b) We establish a novel pipeline by temporal-spatially localizing the scene text region and then decoding them into textual answers. We also enable direct evaluation on the grounded scene-text region.} 
\vspace{-0.1in}
\label{fig:fig1}
\end{figure}

% %\vspace{-0.1in}
\section{Introduction}
\IEEEPARstart{T}{ext}-based Video Question Answering (TextVideoQA) \cite{zhao2022towards, tom2023reading, jahagirdar2023watching} is an emerging task that requires models to answer questions pertaining to scene texts in dynamic visual contents. It challenges the models in identifying a very limited portion of crucial scene texts from a large amount of visual content to answer a particular question, in which motion blur and occlusion often result in wrong predictions (refer to the example in Fig.~\ref{fig:fig1} (a)).
Existing methods \cite{jin2021ruart, zhu2023locate, fang2023separate} that perform well on image-based TextVQA \cite{singh2019towards, biten2019scene} struggle to achieve good performances in the video domain \cite{zhao2022towards}. However, the key factors causing performance loss remain unclear due to the opaque decision-making process. For example, is it because of poor QA or poor scene-text recognition in the video? Additionally, even for the correct predictions, these methods rarely tell if their answers are originated from relevant scene texts in the videos, or attributed to other short-cut paths. This severely impedes further improvements.

To tackle this problem, we propose to study a novel \emph{Grounded TextVideoQA} task by requiring models to answer the question and localize the relevant scene-text regions as evidence. Such a setting enjoys three-fold advantages: \textbf{First}, the grounded scene-text regions serve as visual evidence to support textual answers, thus enabling a reliable TextVideoQA system. 
\textbf{Second}, the grounded scene-text regions can serve as visual answers in scenarios where the accuracy metric fails to align with human intuition to effectively measure the predictions' correctness. For instance, the answer ``30'' in Fig.~\ref{fig:fig1} (a) should be an acceptable (understandable) answer to human. \textbf{Third}, by allowing both visual answers and text answers, we decouple scene-text recognition from VideoQA, thus enabling better diagnose the source of the problem if the models arrive at a wrong prediction. This significantly benefits model debugging towards improvements.
 
To solve \emph{Grounded TextVideoQA}, precisely grounding (localizing) the question-relevant scene texts is a key challenge, especially in both temporal and spatial dimensions under weak supervision. To tackle this, we propose a \emph{Temporal-to-Spatial (T2S)} grounding and then \emph{Question-Answering (QA)} model \model. \model~highlights a temporally- and spatially- disentangled contrastive learning strategy for weakly-supervised scene text grounding in space and time. Specifically, at the first stage, \model~employs temporal grounding to distinguish positive frames (\ie, frames with question-relevant scene texts) from negative frames (\ie, frames without question-relevant scene texts) in a video. {Given that motion blur and occlusion often obscure scene text in video frames, it subsequently refines the selection by identifying a few key positive frames that are most relevant to the question as the grounded frames.} At the second stage, \model~applies spatial grounding in each grounded frame to differentiate positive scene texts from negative ones. {Since answers typically pertain to a small subset of the rich scene text present in a video frame, \model~further selects partial positive scene texts covering the answer as the final grounding results.} Finally, to decode the textual answers from the grounded scene texts, we feed the grounded scene texts, the grounded frames (serve as contexts), and the question into a transformer-based answer decoder for answer prediction. 

As there is a lack of a benchmark for evaluating Grounded TextVideoQA, we construct the \dataset~dataset that considers both answer grounding and QA. \dataset~is built over the existing largest TextVideoQA dataset M4-ViteVQA \cite{zhao2022towards} by extending its QAs in the validation and test sets with spatio-temporal labels (\ie, timestamps and spatial bounding boxes). \dataset ~consists of 2,055 QAs, 729 videos, 2,227 annotated temporal segments, and 52,494 bounding boxes. The labels are manually annotated and checked to be key for deriving accurate textual and visual answers.

Based on our \dataset~dataset, we source the baseline methods from three groups: TextVQA \cite{hu2020iterative}, TextVideoQA \cite{zhao2022towards}, and weakly-grounded VideoQA \cite{gao2023mist,li2023discovering}. {For methods from image-based TextVQA and TextVideoQA, we conduct a post-hoc analysis of the attention distribution corresponding to the predictions}.
For methods designed for weakly-grounded VideoQA, we analyze their grounding outputs derived from the respective grounding modules.
Our findings reveal that existing models struggle to predict visually grounded answers, despite their relatively decent QA behavior. The results show our \model~improves over these methods for both grounding and QA. Additionally, we evaluate two advanced multimodal large language models (MLLMs) on \dataset: the open-source Qwen2-VL~\cite{wang2024qwen2} and the closed-source GPT-4o-mini~\cite{gpt4omini2024}. The experimental results show that T2S-QA has better scene text grounding capabilities than them, indicating the huge room for improvement of current MLLMs in scene-text grounding.
Moreover, we find the performance gap between our \model~and human is still large. For example, humans can correctly ground 77\% of the answers to the questions, whereas \model~only achieves 28\%. In addition, we find that there is a clear discrepancy between grounding and QA performance for both human (77\% \vs ~64\%) and \model~(28\% \vs ~10\%). 
On the one hand, the higher grounding and much lower QA accuracy of \model~indicate that incorrect scene text recognition still seriously jeopardizes the final QA performance. On the other hand, the lower QA accuracy of human confirms the ineffectiveness of exact word matching for answer evaluation, and suggests the advantage and importance of our grounding evaluation to directly accept scene texts as visual answers. Finally, the large discrepancy between \model~and human reveals that the poor TextVideoQA results of existing methods are largely attributed to the challenge posed by scene-text detection and recognition in the video domain.

Our main contributions are summarized as follows:
\begin{itemize}
\item We propose to study a new Grounded TextVideoQA task. It promotes research towards interpretable QA systems by not only providing visual evidence to answers, but also decoupling VideoQA from scene text recognition for better model analysis and development.

\item We propose a cascaded temporal-to-spatial scene-text grounding and question-answering model, \model, with a disentangled temporal-spatial contrastive learning strategy for weakly-supervised scene-text grounding.

\item We manually construct the \dataset~dataset to assess the models' capabilities of scene-text answer grounding.

\item We thoroughly analyze the model behaviors, revealing significant challenges in performing Grounded TextVideoQA.

\end{itemize}

%------------------------------------------------------
\begin{figure*}
\centering
\includegraphics[width=\textwidth]{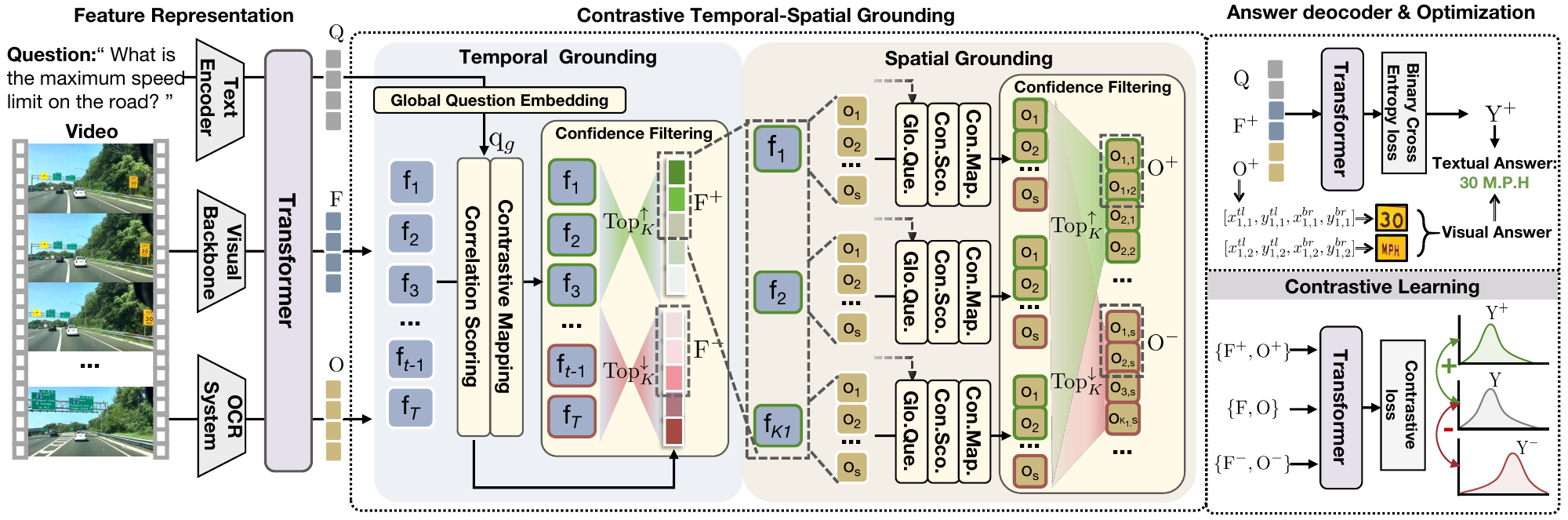}
% \vspace{-0.2in}
\caption{{Overview of our \model~model.} It mainly consists of three components: (1) the Feature Representation prepares the features of the question $\mathrm{Q}$, video frames $\mathrm{F}$, and OCR tokens $\mathrm{O}$; (2) the Contrastive Temporal-Spatial Grounding adopts a two-stage fine-grained grounding approach; (3) the Answer Decoder integrates the grounded frames $\mathrm{F}^{+}$, the grounded OCR tokens $\mathrm{O}^{+}$, and the question $\mathrm{Q}$ to achieve answer generation.
In the Optimization process, we introduce a contrastive learning mechanism to help improve the question answering and answer grounding capabilities of our model.
}
\vspace{-0.1in}
\label{fig:fig2}
\end{figure*}

%------------------------------------------------------

% \vspace{-0.15in}
\section{Related Work}
\label{sec:rel}
\subsection{Text-Based Video Question Answering}
TextVideoQA aims to answer questions by jointly reasoning about scene texts and visual contents in videos. Unlike TextVQA \cite{hu2020iterative, zhou2023exploring, guo2024benchmarking, zhou2024graph}, TextVideoQA \cite{zhao2022towards, tom2023reading, jahagirdar2023watching, zhou2025egotextvqa} focuses on developing the ability of machines to answer questions based on scene texts in dynamic environments. Zhao et al. \cite{zhao2022towards} first identify this challenge and build a benchmark M4-ViteVQA, whose videos factor nine scenarios  (${\textit{i.e.}}$, shopping, traveling, driving, vlog, sport, advertisement, movie, game, and talking), and develop a baseline model T5-ViteVQA. Tom et al. \cite{tom2023reading} propose the RoadTextVQA dataset on driving videos, focusing on questions that require understanding scene texts on road signs. Jahagirdar \etal \cite{jahagirdar2023watching} introduce NewsVideoQA to study TextVideoQA in news videos. 

The above studies have evaluated the QA ability of existing vision-and-language models on TextVideoQA, and shown that there is still large room for improvements. However, these works primarily emphasize the accuracy of the predicted textual answers, overlooking the research on answer interpretability. In this study, we propose Grounded TextVideoQA to explore both answer grounding and question answering. Moreover, we propose a grounding-then-answering method to benchmark the task and help analyze the interplay between scene-text recognition and VQA in videos.

% \vspace{-0.2in}
\subsection{Answer Grounding in VQA.}
A key challenge faced by the VQA \cite{shen2023mutual, Ding2022MuKEAMK, qian2022scene, li2023discovering} community is the issue of language bias \cite{kv2020reducing, song2024emotional, goyal2017making, niu2021counterfactual}. Specifically, models can answer questions by exploiting language priors in datasets without considering the visual content. To some extent, this problem is difficult to detect as commonly used evaluation metrics solely take account into textual answers to assess model performance.
In image-based VQA tasks, many works \cite{chen2022grounding, chen2023vqa, rao2021first, Gan2017VQSLS} propose new benchmarks for answer grounding to support models to localize visual evidence that humans rely on to answer visual questions. This is due to the observation that answer grounding can serve as a valuable foundation for debugging VQA models. However, related research is relatively scarce in VideoQA.
TVQA \cite{lei2018tvqa} and TVQA\texttt{+} \cite{lei2020tvqa+} pioneer in study grounded VideoQA, yet in full supervision. 
NExT-GQA \cite{xiao2023can} firstly studies weakly-supervised grounding for VideoQA, yet focusing on temporal dimension. 
Moreover, all of them concentrate on grounding the normal visual objects or video moments.
In contrast, we focus on spatio-temporally grounding the scene text regions, with our techniques also centering on weakly-supervised grounding. 

\subsection{Weakly-Supervised Video Grounding for VideoQA.}
In VideoQA task \cite{yang2024robust, xiao2021next, wang2021dualvgr, cheng2023keyword}, several methods \cite{qian2023locate, xiao2020visual,xiao2023can, yu2024self, gao2023mist, li2023discovering} introduce weakly-supervised video grounding techniques to improve the QA performance and model interpretability \cite{zhu2020multimedia}. These methods locate critical frames \cite{qian2023locate, li2022invariant, li2023transformer, yu2024self} and visual objects \cite{gao2023mist, li2023discovering,xiao2023contrastive} to aid question answering. For example, \cite{qian2023locate} and \cite{yu2024self} select the temporal proposal with the highest similarity to the language query. \cite{xiao2023can} design a Gaussian mask learning module to ground temporal segments for answer prediction. 

Compared with VideoQA, TextVideoQA focuses on understanding local and even tiny scene texts in the video, where frame-level grounding struggles to capture subtle scene text details. In this work, we study spatio-temporal grounding \cite{zhang2024learning} for the proposed Grounded TextVideoQA task. For video spatio-temporal grounding, MIST \cite{gao2023mist} designs an iterative attention module to select video segment and region features and fuse them for question answering. TranSTR \cite{li2023discovering} locates key video frames and objects and uses object-enhanced frame features for QA. However, these methods rely on the similarity between frame (object) features and questions to select critical elements, lacking effective self-supervision to constrain the selection during training. To address this, inspired by disentangled learning approaches \cite{liu2022disentangled} and contrastive learning strategy \cite{liu2024deep, gao2024guess, gao2025jointly}, we design a disentangled temporal and spatial contrastive learning strategy that distinguishes positive and negative video frames and scene text regions in the video and enlarges the semantic difference between positive and negative features. This approach injects strong self-supervision signals into the model to distinguish question-related video frames and scene texts, thereby effectively achieving scene text grounding.  
Our method distinguishes positive and negative video frames and scene text regions in the video and enlarges the semantic difference between positive and negative features. This approach injects strong self-supervision signals into the model to distinguish question-related video frames and scene texts, thereby effectively achieving scene text grounding.

% \vspace{-0.10in}
\section{Method}
\label{sec:bas}

Given a video and a question, Grounded TextVideoQA aims to encapsulate visual content, scene text, question semantics, and generate textual answers. Meanwhile, the scene text should be explicitly localized to substantiate the predicted textual answer or directly act as  visual answer. To solve the task, localizing the question-related scene text is of crucial importance.
% \textcolor{blue}{As depicted in Fig. \ref{fig:fig2}, 
As such, we propose a cascaded Temporal-to-Spatial Grounding and QA (\model) architecture as depicted in Fig. \ref{fig:fig2}. \model~  comprises three core modules: 1) Feature Representation module extracts features for the question, video frames, and scene texts; 2) the Contrastive Temporal-Spatial Grounding (TSG) module executes temporal grounding to identify question-relevant video frames, followed by spatial grounding to pinpoint critical scene text regions within the identified frames; 3) the Answer Decoder module integrates the grounded video frames, scene text, and question to decode the final answer. For optimization, we 
% Finally, driven by a contrastive learning objective, the Optimization module 
curate positive and negative pairs to optimize the alignment between scene texts and the question spatio-temporally.

% \vspace{-0.15in}
\subsection{Feature Representation}
\label{sec:fea_Rep}
\noindent\textbf{Question features.}
Given a  question, we extract its feature $Q = \{q_l\}^{L}_{l=1}\in\mathbb{R}^{L\times d}$ by a language backbone, where $L$ is the question length, and $d$ is the dimension of word embedding.

\noindent\textbf{Video features.} Given a video, we uniformly sample $T$ frames $V = \{v_{t}\}^T_{v=1}$ and extract frame feature $F = \{f_{t}\}^T_{t=1}\in\mathbb{R}^{T\times d}$ to represent a video. For the $t$-th video frame, its features include two parts: 1) visual feature $f^v_{t} \in \mathbb{R}^{1024}$ extracted by a pre-trained visual backbone which is frozen during training; 2) temporal id feature $f^{te}_{t} \in \mathbb{R}^{50}$ where the temporal id $t$ is assigned to denote the $t$-th video frame and encoded by a embedding layer as a temporal information. We merge two features to obtain the final video frame feature.
\begin{equation}
f_{i}  = \mathrm{LN}({W_{1}}f^v_t + {W_{2}}f^{te}_t ), \label{eq:eq1} 
\end{equation}
where $\mathrm{LN}(\cdot)$ denotes layer normalization, ${W_{1}}$ and ${W_{2}}$ are learnable parameters.

\noindent\textbf{OCR features.} Given the sampled video frames $V = \{v_{t}\}^T_{t=1}$, following \cite{zhao2022towards}, we employ a scene text detection system \cite{wu2021bilingual} to obtain bounding boxes and track ids of the scene texts in each video frame, and then use a scene text recognition system \cite{fang2021read} to read the scene texts according to their bounding boxes. For simplicity, the scene text detection system and recognition system are collectively referred to as optical character recognition (OCR) systems. Finally, the OCR token set for a video is denoted as:
\begin{gather}
\{\text{B}_t, \text{Ta}_t\}^{T}_{t=1} = \text{Detector}(v_t),    \label{eq:eq5} \tag{2} \\
\{r_{t,s}\}^{S}_{s=1} = \text{Recognizer}(v_t, B_t),  \label{eq:eq6} \tag{3} \\
o_{t,s} = \{r_{t,s}, o^b_{t,s}, o^{ta}_{t,s}\}^{T, S}_{t=1, s=1}, \label{eq:eq6} \tag{4}
\end{gather}
where $\text{Detector}(\cdot)$ is the scene text detection system applied to frame $v_t$. $\text{Recognizer}(\cdot)$ is the scene text recognition system applied to frame $v_t$ and its bounding boxes ${\text{B}_t}$, where $B_t = \{o^b_{t,1}, o^b_{t,2}, \dots, o^b_{t,S}\}$ is the set of bounding boxes in frame $t$ and $S$ is the OCR token number in each frame, where each box contains the coordinates $[x_{t,s}^{tl}, y_{t,s}^{tl}, x_{t,s}^{br}, y_{t,s}^{br}]$ for the top-left and bottom-right corners. $\text{Ta}_t = \{o^{ta}_{t,1}, o^{ta}_{t,2}, \dots, o^{ta}_{t,S}\}$ is set of track ids for the detected text regions in frame $t$. $r_{t,s}$ is the recognized scene text content.

After that, we obtain OCR features $O = \{o_{t, s}\}^{T, S}_{t=1, s=1}\in\mathbb{R}^{T \times S \times d}$ in the video. For the $s$-th OCR token in the $t$-th frame, its features consist of five aspects: 1) FastText feature $o^f_{t,s}\in\mathbb{R}^{300}$ \cite{bojanowski2017enriching} providing essential word information based on the recognized scene text content $r_{t,s}$; 2) pyramidal histogram of characters (PHOC) feature $o^p_{t,s}\in\mathbb{R}^{604}$ \cite{almazan2014word} representing character features based on the recognized scene text content $r_{t,s}$; 3) normalized bounding box feature $o^b_{t,s}\in\mathbb{R}^{4}$ as the visual geometric features; 4) temporal id feature $o^{te}_{t,s}\in\mathbb{R}^{50}$ to provides temporal features consistent with the video frames; 5) track id feature $o^{ta}_{t,s}\in\mathbb{R}^{50}$ as the identification features.  To project the representations into a common $d$-dimensional space, we apply a linear mapping on the above features respectively and sum them up to get the final representation of the OCR token features. 
% \vspace{-0.05in}
\begin{align}
o_{t,s}\!&=\!\mathrm{LN}({W_{3}}o^f_{t,s}\!+\!{W_{4}}o^p_{t,s}\!+\!{W_{5}}o^{te}_{t,s}\!+\!{W_{6}}o^{ta}_{t,s} )\!+\!\mathrm{LN}({W_{7}}o^b_{t,s}), \label{eq:eq2} \tag{5}
\end{align}
where $\mathrm{LN}(\cdot)$ denotes layer normalization, ${W_{3}}\!\sim\!{W_{7}}$ are learnable parameters.

To reduce the semantic gap between different modalities, we apply a two-layer transformer $\varPsi$ \cite{vaswani2017attention} to establish associations between visual and textual content, aiding in the update of features. Here, we obtain the new question feature $\mathrm{Q} = \{\mathrm{q}_l\}^L_{l=1}$, frame feature $\mathrm{F} = \{\mathrm{f}_t\}^T_{t=1}$, and OCR token feature $\mathrm{O} = \{\mathrm{o}_{t, s}\}^{T, S}_{t=1, s=1}$. 
% \vspace{-0.05in}
\begin{align}
 [\mathrm{Q}; \mathrm{F}; \mathrm{O}] &=\varPsi([Q; F; O]), \label{eq:eq3} \tag{6}
\end{align}

% \vspace{-0.25in}
\subsection{Contrastive Temporal-Spatial Grounding}
Accurately localizing question-relevant scene text in a video is challenging, especially under weak supervision. To address this problem, we design an adaptive two-stage approach to localize key video frames and OCR tokens for answering questions: a Temporal Grounding (TG) stage that identifies the question-relevant video frames and a Spatial Grounding (SG) stage that pinpoints the critical OCR tokens.

\subsubsection{\textbf{Temporal Grounding}}
In most cases, scene text relevant to questions often appears in specific video frames as shown in Fig.~\ref{fig:fig4}. To align video content with question semantics and ground key video frames as context for effective question answering, we design a multi-step approach as follows.

\noindent \textbf{Global Question Embedding.} A global sentence representation, $\mathrm{q}_g$, is derived from word embeddings $\mathrm{q}_l$ using self-attention, capturing overall question semantics by emphasizing keywords. 
\begin{gather}
\mathrm{q}_g = \sum_{l=1}^{L} \mathrm{Softmax}(W_{8}\mathrm{q}_l) \cdot \mathrm{q}_l, \label{eq:eq4} \tag{7}
\end{gather}

{\noindent \textbf{Correlation Scoring.} The correlation-based softmax operation computes the relevance distribution between the global question embedding and the video frame features. Guided by the contrastive learning objective in Eqn. \eqref{eq:eq11}, we calculate the cross-attention scores for the positive frame distribution $\mathrm{f}^{+}\in\mathbb{R}^{T \times 1}$ and the negative frame distribution $\mathrm{f}^{-}\in\mathbb{R}^{T \times 1}$, respectively. This approach explicitly distinguishes between positive (relevant) and negative (irrelevant) frames, ensuring that the positive frames align with the question semantics. The contrastive learning objective further amplifies the semantic differences between positive and negative frames, enhancing the capture of question-relevant video information.}
\begin{gather}
\mathrm{f}^{+} = \mathrm{Softmax}({W_9}\mathrm{F} \cdot {W_{10}}\mathrm{q}_g^\top), \label{eq:eq5} \tag{8} \\
\mathrm{f}^{-} = \mathrm{Softmax}({W_{11}}\mathrm{F} \cdot {W_{12}}\mathrm{q}_g^\top), \label{eq:eq6} \tag{9}
\end{gather}

\noindent \textbf{Contrastive Mapping.} To identify video frames that are relevant or irrelevant to the question, we apply the Gumbel-Softmax function to obtain the distribution of positive ($\mathrm{f}^+$) and negative ($\mathrm{f}^-$) frames. This operation enables a discrete distinction, ensuring that the positive frames align with the question semantics.
\begin{gather}
\mathrm{M}_f = \mathrm{Gumbel\text{-}Softmax}([\mathrm{f}^{+}; \mathrm{f}^{-}]), \label{eq:eq7} \tag{10}
\end{gather}
where [;] denotes concatenation, where $W_{8}\!\sim\!W_{12}$ are learnable parameters. The first and second columns of $\mathrm{M}_f$ index the positive frames (${\textit{i.e.}}$, $\mathrm{M}_{f_{0}}$) and negative frames (${\textit{i.e.}}$, $\mathrm{M}_{f_{1}}$).

\noindent \textbf{Frame Filtering.} 
Not all positive video frames carry clear and unobstructed scene text to answer the question. We need to further filter the positive and negative video frames based on their relevance to the given question. Specifically, we utilize Eqns \eqref{eq:eq8} and \eqref{eq:eq9} to enhance the semantic distinction between positive and negative frame features. Leveraging the cross-correlation scores computed by Eqns \eqref{eq:eq5} and \eqref{eq:eq6}, we identify the top $K_1$ video frames from the positive (negative) frame set that exhibit the highest (lowest) relevance to the question. The obtained frame sets \( \{\mathrm{F}^+, \mathrm{F}^-\} \) demonstrate significant semantic divergence, quantified by their relevance to the question. Guided by the contrastive learning objective, the model is optimized to further amplify this divergence, enabling grounding of question-relevant video frames \( \mathrm{F}^+ \).
\begin{align}
\mathrm{{F}^{+}} &= \textrm{Top}_K^{\uparrow}(\mathrm{f}^{+}, \mathrm{M}_{f_{0}}\mathrm{F}) |_{K = K_1},  \label{eq:eq8} \tag{11} \\
\mathrm{{F}^{-}} &= \textrm{Top}_K^{\downarrow}(\mathrm{f}^{-}, \mathrm{M}_{f_{1}}\mathrm{F}) |_{K = K_1}, \label{eq:eq9} \tag{12} 
\end{align}

Up to now, we have introduced how to ground the frames \( \mathrm{F}^+ \) that carry the key scene text for answering questions. These frames will be inputted into the answer decoder (Sec.~\ref{sec:ans}) to serve as context for answer prediction.

\subsubsection{\textbf{Spatial Grounding}}
Our preliminary analysis reveals that most answers only occupy a very small area in the video frame, as shown in Fig.~\ref{fig:fig4}. Therefore, we first need to distinguish the scene texts related to the question from numerous irrelevant visual elements, and then locate a few critical strongly associated scene texts as visual answers. 
To achieve that, SG facilitates adaptive OCR token selection similar to Eqn. \eqref{eq:eq4} $\sim$ \eqref{eq:eq9}. Concretely, for the $i$-th critical frame, SG is fed with $S$ object features detected on the frame, enabling it to differentiate positive OCR tokens from negative ones based on their similarity to the given question. For positive OCR tokens, we select $K_2$ OCR tokens in each frame with the \emph{highest} attention weight from $\mathrm{o}^{+}$ and gather their corresponding OCR features as $\mathrm{{O}^{+}}\in\mathbb{R}^{K_1 \times K_2\times d}$. For negative OCR tokens, we select $K_2$ OCR tokens in each frame with the \emph{lowest} attention weight from $\mathrm{o}^{-}$ and gather their corresponding OCR features as $\mathrm{{O}^{-}}\in\mathbb{R}^{K_1 \times K_2\times d}$. It is worth noting that the detected OCR results vary for each frame, and SG is applied independently to each frame. Hence, even if we use hyperparameter $K_2$ to select OCR tokens in each frame, the numbers of critical OCR tokens in each frame are different.

Up to now, we have obtained the pair of the $K_1$ positive video frames and $K_1 \times K_2$ positive OCR tokens, namely $\{\mathrm{{F}^{+}}, \mathrm{{O}^{+}}\}$, and the pair of the $K_1$ negative video frames and $K_1 \times K_2$ negative OCR tokens, namely $\{\mathrm{{F}^{-}}, \mathrm{{O}^{-}}\}$. Notably, we take the bounding boxes of $\mathrm{{O}^{+}}$ as grounded OCR tokens for answer grounding evaluation, the evaluation details are elaborated in Sec. \ref{expr:eval}.

% \vspace{-0.15in}
\subsection{Answer Decoder and Optimization}
\label{method:ans}
To utilize the grounded video frames and OCR tokens to generate answers, we adopt a contrastive learning mechanism as an auxiliary objective to enhance the grounding accuracy. Specifically, we obtain the positive frame-OCR token pair $\{\mathrm{{F}^{+}}, \mathrm{{O}^{+}}\}$ and the negative frame-OCR token pair $\{\mathrm{{F}^{-}}, \mathrm{{O}^{-}}\}$ in TSG, and take raw video frames and OCR tokens $\{\mathrm{{F}}, \mathrm{{O}}\}$ as anchor. We next feed these pair instances into the answer decoder to obtain three answer predictions, ${\textit{i.e.}}$, anchor $\mathrm{Y}$ and its contrastive counterparts $\mathrm{Y}^{+}$ and $\mathrm{Y}^{-}$. We elaborate on the answer generation and optimization process below.

%------------------------------------------------------

\subsubsection{\textbf{Answer Decoder}}
\label{sec:ans}
Following \cite{hu2020iterative}, the answer decoder is composed of a three-layer transformer $\hat \varPsi$ and two classifiers --- a vocabulary classifier $\psi_{voc}$ and an OCR token classifier  $\psi_{ocr}$. 
Take the example of the answer $\mathrm{Y}^{+}$ generated by the positive frame-OCR token pair $\{\mathrm{{F}^{+}}, \mathrm{{O}^{+}}\}$ ,
we concatenate the question $\mathrm{Q}$, the positive frames $\mathrm{{F}^{+}}$ and the positive OCR tokens $\mathrm{{O}^{+}}$, and a hidden state $\mathrm{D}$ and input them into the transformer block as follows:
%\vspace{-0.1in}
\begin{align}
[\mathrm{Q}_{t\!+\!1};\!\mathrm{{F}^+_{t\!+\!1}};\!\mathrm{{O}^+_{t\!+\!1}};\!\mathrm{D}_{t\!+\!1}]\!=\!\hat \varPsi([{W_{13}}\mathrm{Q_t};\!{W_{14}}\mathrm{{F}^{+}_t};\!{W_{15}}\mathrm{{O}^{+}}_t;\!{W_{16}}\mathrm{D}_t]),\!\label{eq:eq10} \tag{13} 
\end{align}
where $W_{13}\!\sim\!W_{16}$ are learnable parameters, 
the hidden state $\mathrm{D}_0$ is initialized by the positional embedding \cite{hu2020iterative}, and where the initialization of $\mathrm{F^+_0}=\mathrm{F^+}$, $\mathrm{O^+_0}=\mathrm{O^+}$, and $\mathrm{Q_0}=\mathrm{Q}$. By iteratively performing the transformer ${L_{a}}$ times, we obtain $\mathrm{D}\!=\![d_1$, $ \cdots$, $d_{L_{a}}]\!\in\!\mathbb{R}^{{L}_{a}\times d}$. 
At the $t$-th decoding step, the classifier $\psi_{voc}$ is optimized to predict the probability score $y^{voc}_t$ over a preset vocabulary. The classifier $\psi_{ocr}$ calculates the relevance score $y^{ocr}_t$ of $\mathrm{D}_t$ and the OCR token $\mathrm{{O}^{+}}$. 
We implement the $\textrm{Argmax}$ function on $y^{voc}_t$ and $y^{ocr}_t$ to predict the word $y_t$. Thus, the generated answer is denoted as $\mathrm{Y}^{+} = \{y_1, \cdots, y_{L_{a}}\}$. The generation of the negative answer $\mathrm{Y}^{-}$ and the anchor answer $\mathrm{Y}$ is the same as above.

\subsubsection{\textbf{Model Optimization}}
\label{sec:opti}
By far, we obtain the three answers $\mathrm{Y}$, $\mathrm{Y}^{+}$, and $\mathrm{Y}^{-}$ based on the anchor, positive frame-OCR token pair, and negative frame-OCR token pair. Next, we optimize the model from two aspects: 1) we design a contrastive loss to improve the grounding accuracy of our model; 2) we adopt a cross-entropy loss to help the model utilize the localized features to answer questions.

\noindent\textbf{Contrastive loss.}
To enhance the grounding accuracy of the Grounded TextVideoQA model, we exploit the causal invariance between the original video and the positive video frames. The contrastive learning loss is designed to amplify the semantic differences between the positive frames (OCR tokens) and the negative frames (OCR tokens). We construct a contrastive objective referring to InfoNCE \cite{oord2018representation} as follows:
%\vspace{-0.05in}
\begin{gather}
\mathcal{L}_{cons}\!=\!-\!\log\!\left(\!\frac{\exp(\mathrm{sim}(\mathrm{Y}^{+}\!,\!\mathrm{Y}^\top)\!/\!\tau)}{\exp(\mathrm{sim}(\mathrm{Y}^{+}\!,\!\mathrm{Y}^\top)\!/\!\tau)\!+\!\exp(\mathrm{sim}(\mathrm{Y}^{-}\!,\!\mathrm{Y}^\top)\!/\!\tau)}\!\right)\!, \label{eq:eq11} \tag{14}
\end{gather}
where $\mathrm{sim}(\cdot)$ denotes cosine similarity function, and {we empirically set temperature coefficient $\tau$ = 0.1 following~\cite{chen2020simple, li2022equivariant}.}

\noindent\textbf{Binary cross-entropy loss.}
For TextVideoQA, the answer prediction can be regarded as a multi-label classification problem with the classifier $\psi_{voc}$ and the classifier $\psi_{ocr}$. {Binary cross-entropy loss function} is suitable for multi-label classification \cite{hu2020iterative, zhao2022towards}. Considering that the grounded positive video frames and OCR tokens are beneficial for question answering, we utilize this loss function to minimize the semantic distance between the positive answer and the ground truth answer.
\begin{align}
\mathcal{L}_{bce} = - {\mathrm{Y}} \log (\sigma({\mathrm{Y^{+}}})) - (1 - {\mathrm{Y}}) \log (1 - \sigma (\mathrm{Y^{+}})), \label{eq:eq12} \tag{15} 
\end{align}
where  $\sigma$($\cdot$) is sigmiod function, $ {\mathrm{Y}}$ is the ground truth answer.

The overall training objective of \model~is as follows:
\begin{align}
\mathcal L &= \mathcal L_{bce} + \lambda  \mathcal L_{cons},  \label{eq:eq13} \tag{16} 
\end{align}
and $\lambda$ is a trade-off hyperparameter.

%------------------------------------------------------
\begin{figure}[t]
\centering
\includegraphics[width=\columnwidth]{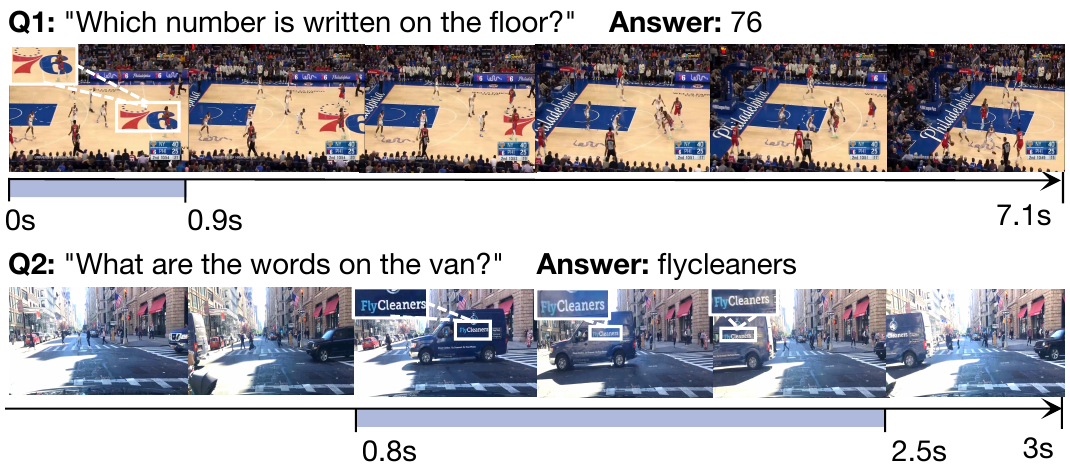}
% %\vspace{-0.25in}
\caption{Examples of spatial-temporal labels in \dataset.}
\label{fig:fig4}
%\vspace{-0.1in}
\end{figure}

\begin{table}[t]
\caption{Statistics of \dataset~ dataset.}
%\vspace{-0.08in}
\label{tab:tab2}
\setlength{\tabcolsep}{.7em}
% \fontsize{7}{9}\selectfont
\resizebox{\columnwidth}{!}{
\begin{threeparttable}
\begin{tabular}{l|cccccc}
\Xhline{1pt}
\makecell[l]{Split} & \makecell[c]{Vid.} & \makecell[c]{Que.} &  \makecell[c]{Tem. Span} & \makecell[c]{Box Num.} & \makecell[c]{Ave. Box}   \\
\Xhline{1pt}
Train & 5,444 & 18,220 & -  & -  & -\\ 
Val  & 343 &  981  & 1,087  & 25,445 &  25.88 \\ 
Test & 386 &  1,074 & 1,140 & 27,049 &  25.19 \\  
\Xhline{1pt}
\end{tabular}
\end{threeparttable}}
\vspace{-0.15in}
\end{table}

% %\vspace{-0.15in}
\section{Dataset}
\label{sec:dat}
\subsection{\dataset~ Construction}
\noindent\textbf{Data Source.} We choose M4-ViteVQA \cite{zhao2022towards} dataset as our data source to augment with spatio-temporal labels. 
Compared with others (Table \ref{tab:tab1}), M4-ViteVQA has a richer set of video scenarios, such as shopping, traveling, driving, vlog, sport, advertisement, movie, game, and talking. Specifically, we choose its hard data split where video clips do not overlap among the train/val/test set. The data split consists of 5,444/525/680 videos and 18,220/1,620/2,103 questions on train/val/test sets. 

\noindent\textbf{Dataset Filtering.} We only label the validation and test sets for a weakly-supervised setting. {To ensure the clarity of answer grounding,
the questions whose answers do not involve scene texts or involve multiple text regions on a single frame are removed before the annotation}. Moreover, the questions whose answer texts are blurry or occluded in the video frame are also removed. Consequently, 343/386 videos and 981/1,074 questions on val/test sets are to be annotated.

\noindent\textbf{Label Collection.} We use Vidat \cite{zhang2020vidat} as our video annotation tool, and invite 15 undergraduate students for annotation. The annotators are instructed to first comprehend the questions and video contents, and then annotate scene texts pertaining to the answers in the video with bounding boxes. Noteworthy, all frames that carry the answer should be annotated for a given question. Moreover, if an answer comprises multiple words, we ask for a single bounding box that can encompass all answer words. Yet, our preliminary statistics show that the majority of answers include only a single word. Note that there is at most one annotated box on a video frame. Fig. \ref{fig:fig4} illustrates two examples from the ViTXT-GQA dataset, where we need to annotate the temporal segment timestamps for the appearance of the scene text corresponding to the answer for each question. Besides, we annotate a bounding box on each frame, even if the answer consists of multiple words. To ensure quality and reduce subjectivity, each QA is reviewed by at least two people. The annotation process took about two months.

%------------------------------------------------------
\begin{figure}[t]
\centering
\includegraphics[width=\columnwidth]{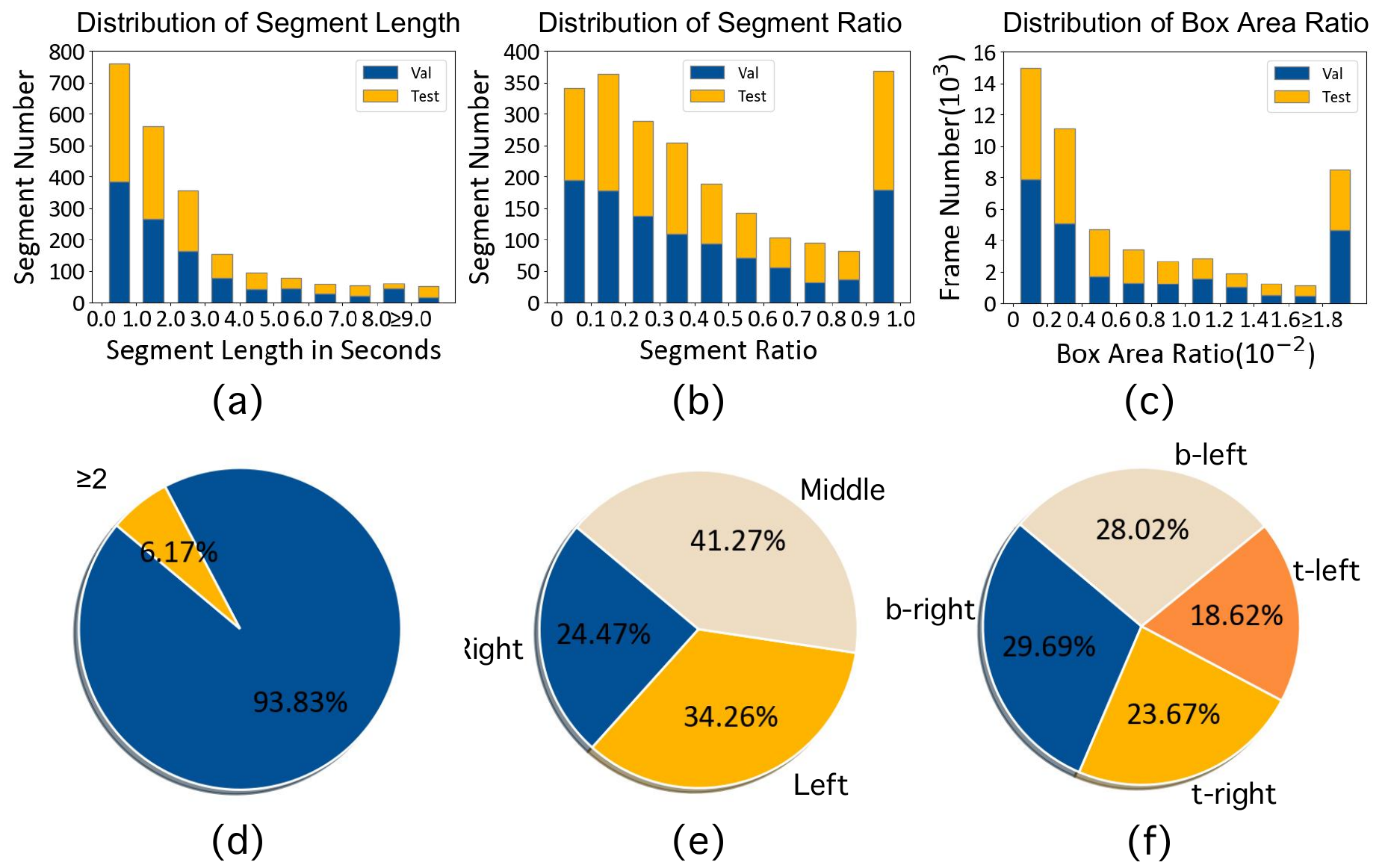}
% %\vspace{-0.25in}
\caption{Analysis of annotated spatio-temporal labels.}
\label{fig:fig3}
\vspace{-0.1in}
\end{figure}

% \vspace{-0.15in}
\subsection{{Dataset Analysis}}
The statistics of \dataset~are shown in Table \ref{tab:tab2} and Fig.~\ref{fig:fig3}. In Table \ref{tab:tab2}, we annotate 1,087 and 1,140 temporal spans on the validation and the test sets, including 25,445 and 27,049 box annotations. The average box annotation per video is about 25. Fig. \ref{fig:fig3} (a) and (b) show the temporal ratio distributions of the annotated temporal segments. In 722/734 questions of the validation/test sets, the annotated temporal segment ratio in the video is less than 50\%; in 123/118 questions of the validation/test sets, the annotated segment ratio is greater than 98\%. In the validation set and test set, the average ratio of annotated segments is 41.75\% and 44.33\% respectively. 
Fig. \ref{fig:fig3} (d) shows that a significant 93.83\% of the questions have one temporal segment. Fig. \ref{fig:fig3} (e) reflects the spatial distribution of the annotated bounding boxes within the temporal segments. The annotated segments are evenly distributed across the left, middle, and right sections of the video, with 41.27\% of the bounding boxes distributed in the middle portion of the video. We also examine the spatial distribution characteristics of the bounding boxes. Fig. \ref{fig:fig3} (f) shows the spatial distribution of the bounding boxes within the image. It can be observed that the boxes are uniformly distributed in the top-left, bottom-left, top-right, and bottom-right corners of the video frames, with a slightly higher occurrence of boxes in the bottom-right corner, reaching 29.69\%. Fig. \ref{fig:fig3} (c) presents the statistics of the ratio between the annotated bounding box area and the area of the corresponding video frame. It is found that the majority of the boxes have an area ratio smaller than {0.2\%}, which reflects the challenge of locating scene text in videos for QA.

\begin{table}[t]
\caption{Benchmark comparison. ST: Scene Text. TG: Temporal Grounding. SG: Spatial Grounding. Acc: Accuracy. IoU: Intersection over Union. GQA: Accuracy \& IoU.}
%\vspace{-0.08in}
\label{tab:tab1}
\fontsize{6}{9}\selectfont
\small
\resizebox{\columnwidth}{!}{
\begin{threeparttable}
\begin{tabular}{l|ccccc}
\Xhline{1pt}
\makecell[l]{Datasets} & \makecell[c]{ST}  & \makecell[c]{TG} & \makecell[c]{SG} & \makecell{Evaluation Metrics}   \\
\Xhline{1pt}
TVQA \cite{lei2018tvqa} & $\times$ & $\checkmark$ & $\times$ & Acc, IoU  \\ 
TVQA\texttt{+} \cite{lei2020tvqa+} & $\times$ & $\checkmark$ & $\checkmark$ & Acc, IoU  \\ 
NExT-GQA \cite{xiao2023can} & $\times$ &  $\checkmark$ & $\times$ & Acc, IoP, GQA  \\
\hline
NewsVideoQA \cite{jahagirdar2023watching} & $\checkmark$ & $\times$  & $\times$ & Acc, ANLS  \\ 
RoadTextVQA \cite{tom2023reading} & $\checkmark$ & $\times$  & $\times$ & Acc, ANLS  \\ 
M4-ViteVQA \cite{zhao2022towards} & $\checkmark$ & $\times$ & $\times$ & Acc, ANLS   \\
\hline
\dataset & $\checkmark$ & $\checkmark$ & $\checkmark$ & Acc, ANLS, IoU, GQA  \\ 
\Xhline{1pt}
\end{tabular}
\end{threeparttable}}
\vspace{-0.15in}
\end{table}

% %\vspace{-0.15in}
\subsection{{Dataset Comparison}}
To highlight the uniqueness of \dataset, we compare it with TextVideoQA benchmarks and visually grounded VideoQA benchmarks in Tab. \ref{tab:tab1}. 

\noindent\textbf{Comparison with TextVideoQA.} NewsVideoQA \cite{jahagirdar2023watching}, RoadTextVQA \cite{tom2023reading}, and M4-ViteVQA \cite{zhao2022towards} aim to predict textual answers. \dataset~ differs in two main aspects: (1) it provides visual evidence to support the answers, and (2) it extends the VQA setting by allowing visual answers. This satisfies more practical applications and facilitates better diagnostics of model performance. For example, are the incorrect answer predictions due to scene text recognition errors or overly strict answer evaluation? \dataset ~is also more challenging because: (1) models need to achieve multiple objectives (${\textit{i.e.}}$, grounding and QA) while maintaining consistency, and (2) models must accomplish not only temporal grounding but also spatial grounding, which is challenging to locate scene text that answers the question in visually rich videos.

\noindent\textbf{Comparison with VideoQA.} 
The fully supervised benchmark TVQA \cite{lei2018tvqa} and TVQA+ \cite{lei2020tvqa+} respectively provides temporal annotations and spatio-temporal annotations in the training data, where labels can resolve referential ambiguities in questions or improve QA performance through well-localized visual content. \dataset~differs from them by focusing on recognizing and grounding \emph{scene text} as visual evidence for QA. Additionally, we place emphasis on weakly supervised spatio-temporal grounding to better investigate model generalization. Compared with NExT-GQA \cite{xiao2023can}, which focuses on weakly supervised temporal grounding, we consider fine-grained spatio-temporal grounding and additionally incorporate scene text understanding, making it more challenging and offering broader applications.

\begin{table*}[t]
\caption{Performance comparison on \dataset~test set. 
Models use ABINet \cite{fang2021read} by default for scene text recognition. We additionally upgrade to CLIPOCR \cite{wang2023symmetrical}. The best and second-best results are \textbf{bolded} and \underline{underlined} respectively.}
\label{tab:tab3}
\setlength{\tabcolsep}{.4em}
\centering
\fontsize{8}{10}\selectfont
\resizebox{\textwidth}{!}{
\begin{threeparttable}
\begin{tabular}{l|cccccc|cccccc}
\Xhline{1pt}
\multirow{2}{*}{\makecell[c]{Method}} & \multicolumn{6}{c|}{Top $1 \times 1$} & \multicolumn{6}{c}{Top $5 \times 5$}  \\
\cline{2-13}
& \makecell[c]{IoU@0.5} & \makecell{IoU@0.3} & \makecell[c]{Acc} & \makecell[c]{{ANLS}} &  \makecell[c]{GQA@0.5} & \makecell{GQA@0.3} & \makecell[c]{IoU@0.5} & \makecell{IoU@0.3} & \makecell[c]{Acc} & \makecell[c]{ANLS}  & \makecell[c]{GQA@0.5} & \makecell{GQA@0.3}   \\
\Xhline{1pt}
M4C~\cite{hu2020iterative} & 2.45 & 4.41 & 9.66 & 0.1362 & 0.64 & 0.70 & 9.13 & 14.35 & 9.87  & 0.1344  & 2.29 & 2.56 \\ 
T5-ViteVQA~\cite{zhao2022towards}  & 1.30  & 2.15 & 9.45  & 0.1381 & 0.36 &  0.45 & 16.82 & 24.81 & 9.45 & 0.1381 & 1.81 & 2.03    \\
MIST~\cite{gao2023mist} & 1.80 & 2.94 & 8.47 & 0.1320 & 0.38 & 0.42 & 10.69  & 16.78 & 8.53 & 0.1316  & 1.72 & 1.18 \\
TranSTR~\cite{li2023discovering} & 2.72 & 4.26 & 9.45 & 0.1302 & 0.45 & 0.45 & 14.94 & 23.01 & 9.27 & 0.1430 & 2.63  &  3.05 \\
\hline
\textbf{\model~(ABINet)}  & \underline{2.99} & \underline{4.89}  & \underline{9.72} & \underline{0.1445} & \underline{0.72} & \underline{0.72} & \underline{28.14} & \underline{41.62} & \underline{10.20} & \underline{0.1472} & \underline{2.69} & \underline{3.62}  \\
\textbf{\model~(CLIPOCR)} & \textbf{3.21} & \textbf{4.95} & \textbf{9.87} & \textbf{0.1477} & \textbf{0.88} & \textbf{0.88} & \textbf{28.60} & \textbf{41.94} & \textbf{10.45} & \textbf{0.1520} & \textbf{3.50}  & \textbf{ 4.23 } \\
\Xhline{1pt}
\end{tabular}
\end{threeparttable}}
\vspace{-0.15in}
\end{table*}

%------------------------------------------------------
\begin{table}[t]
\caption{{Performance of MLLMs (zero-shot) on \dataset. 
The evaluation is conducted under the Top 1$\times$1 setting.}} 
\label{tab:r1_mllm}
\setlength{\tabcolsep}{.3em}
\fontsize{9}{11}\selectfont
\resizebox{\columnwidth}{!}{
\begin{threeparttable}
\begin{tabular}{l|cccc}
\Xhline{1pt}
\makecell[c]{Method} & \makecell[c]{IoU@0.5} & \makecell[c]{Acc} & \makecell[c]{ANLS}  & \makecell[c]{GQA@0.5} \\
\Xhline{1pt}
\multicolumn{5}{l}{\emph{\textcolor{gray}{Question Answering}}} \\
GPT-4o-mini w/ Standard & - & 34.19 & 0.5070 & - \\
Qwen2-VL w/ Standard & -  & \underline{41.34} & \underline{0.5649} & - \\
\hline
\multicolumn{5}{l}{\emph{\textcolor{gray}{Question Answering with CoT Prompt}}} \\
GPT-4o-mini w/ CoT & - & {37.99} &  {0.5574}  & - \\
Qwen2-VL w/ CoT  & - & \textbf{41.90} & \textbf{0.5663} \\
\hline
\multicolumn{5}{l}{\emph{\textcolor{gray}{Question Answering with Grounded CoT Prompt}}} \\
GPT-4o-mini w/ Grounded CoT   & \cellcolor{gray!30}0.19  &  34.46  &  0.5058  &  \cellcolor{gray!30}0.00 \\
Qwen2-VL w/ Grounded CoT & \cellcolor{gray!30}{2.14} & 26.07 & 0.3628 & \cellcolor{gray!30}\underline{1.02} \\ 
Qwen2-VL w/ One-shot Grounded CoT & \cellcolor{gray!30}\underline{{2.94}} & 26.98 & {0.3664} & \cellcolor{gray!30}\textbf{{1.32}}    \\ 
\hline
T2S-QA~(CLIPOCR) & \cellcolor{gray!30} \textbf{3.21} & {9.87} & {0.1477} & \cellcolor{gray!30}{0.88} \\ 
\Xhline{1pt}
\end{tabular}
\end{threeparttable}}
\vspace{-0.15in}
\end{table}

%------------------------------------------------------
\begin{figure}
\centering
\includegraphics[width=\columnwidth]{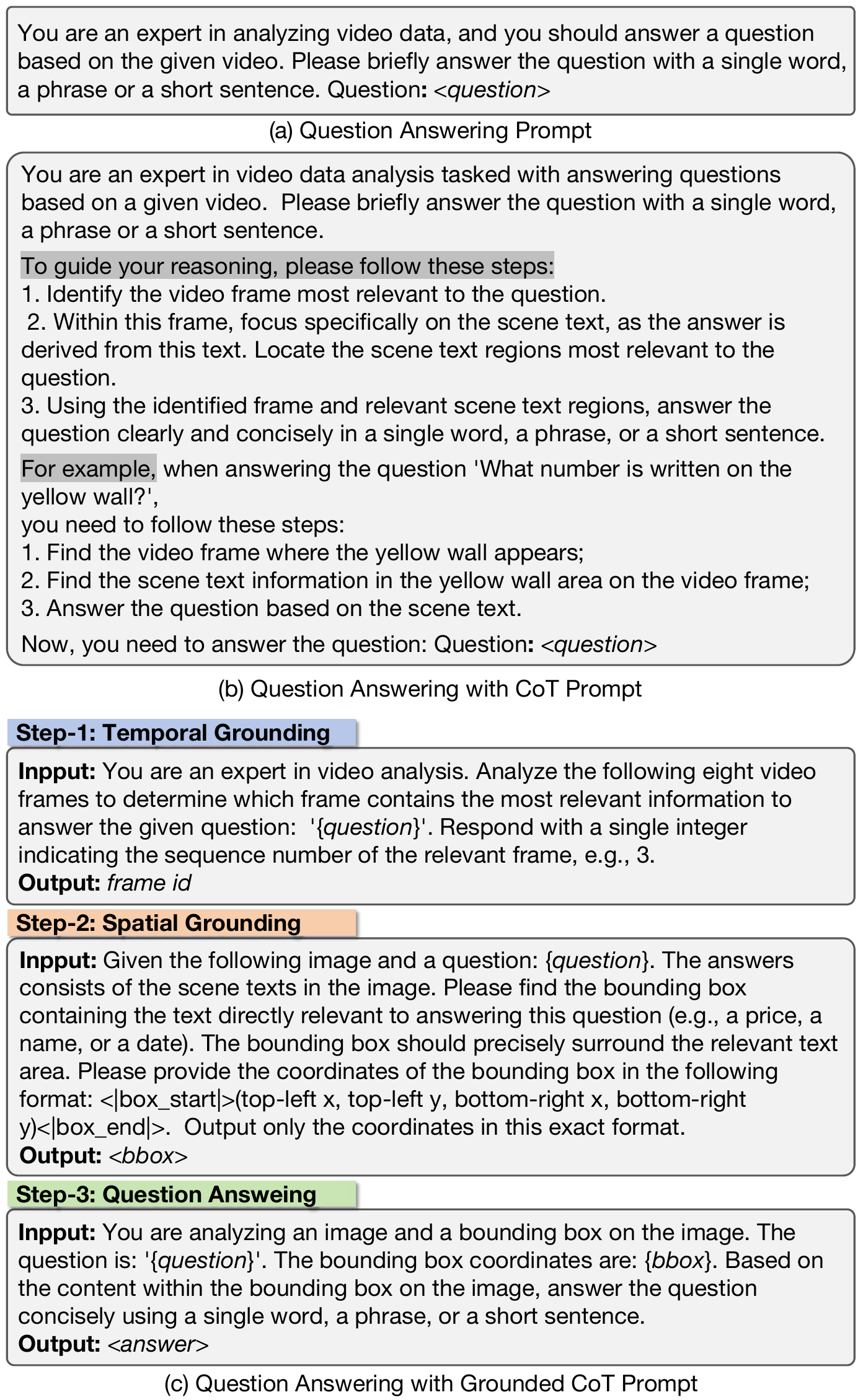}
\caption{
{Prompts for MLLMs to perform ViTXT-GQA. }}
\label{fig:r1}
\vspace{-0.1in}
\end{figure}

\section{Experiments}
\label{sec:exp}
In this section, we show the experimental results to answer three research questions. 

\noindent{$\bullet$} \textbf{Q1:} How effective is our \model ~model?

\noindent{$\bullet$} \textbf{Q2:} What is the contribution of each component in \model?

\noindent{$\bullet$} \textbf{Q3:} How does scene-text recognition affect TextVideoQA?

% \vspace{-0.15in}
\subsection{Settings}
\noindent{\textbf{Implementation Details.}}
We adopt a pre-trained BERT-Base \cite{Devlin2019BERTPO} model to encode the question. The encoder is also finetuned. For video embedding, each frame is encoded by a ViT-L \cite{dosovitskiy2020image} model pre-trained on ImageNet-21k. We standardize the video frame rate to 10 frames per second and set frame number $T$ = 64, OCR token number $S$ = 15 in each frame, the maximal question length $L$ = 20, and the common hidden dimension $d$ = 768. For answer decoder and model optimization, we set the step number of answer decoding to $L_{a}$ = 12 and the trade-off hyperparameter $\lambda$ = 100. We collect the top 5,000 frequent words from the answers in the training set as a fixed vocabulary. During training, we set the batch size to 48  and train for a maximum of 24,000 iterations. Our model is trained using the Adam optimizer, with a learning rate of 1e-4 and a staircase learning rate schedule, where we multiply the learning rate by 0.1 at 10,000 and at 20,000 iterations. 
The best checkpoint is selected using the validation set accuracy.

\noindent{\textbf{Evaluation.}}
\label{expr:eval}
We consider QA and answer grounding evaluation. For QA, we adopt {Accuracy} ({\textbf{Acc}}) in percentage $\in[0, 100]$ and {Average Normalized Levenshtein Similarity} (\textbf{ANLS}) \cite{biten2019scene} $\in[0, 1]$. {Acc} considers a predicted answer to be correct if and only if it matches the ground-truth answer word by word. {ANLS} measures the character-level compositional similarity between the predicted answer and the ground truth. For answer grounding, we adopt Intersection over Union (\textbf{IoU}) between the grounded and the annotated OCR token boxes.
We consider IoU at two thresholds of 0.3 and 0.5 for better analysis.
To evaluate joint QA and answer grounding, we consider grounded QA accuracy (\textbf{GQA}) \cite{xiao2023can}. Specifically, we report GQA@0.5 and GQA@0.3 metrics which stand for the percentages of correctly answered questions with grounding IoU $\geq$ 0.5 and 0.3 respectively.
Finally, we set two evaluations considering the severe challenge of video scene-text grounding. The first is a normal evaluation of the Top 1$\times$1, which measures the Top-1 grounded scene text region within the Top-1 temporally localized video frame. The second is Top 5$\times$5, which relaxes the requirements from Top-1 to Top-5 for both spatial and temporal grounding, \ie, measuring if the identified Top-5 frames and the Top-5 regions within each frame contains the ground-truth scene-text region.

\subsection{{Main Comparison (RQ1)}}
As there is no existing method for Grounded TextVideoQA. We adapt several approaches from TextVQA, TextVideoQA, and VideoQA as baselines in Tab.~\ref{tab:tab3}. For a fair comparison, all methods in Tab.~\ref{tab:tab3} share feature encoding and answer decoding backbones. Besides, {we evaluate the zero-shot performance of the current advanced MLLMs Qwen2-VL~\cite{wang2024qwen2} and GPT-4o-mini~\cite{gpt4omini2024} in Tab.~\ref{tab:r1_mllm}.}  We elaborate on these model details as follows.

\textbf{M4C} \cite{hu2020iterative} is an image-based TextVQA model. We adapt it to the video domain by randomly sampling a single frame. In practice, we sample the first, middle, and end frames, and average their results for reference.
To obtain spatial grounding results, we employ post-hoc attention analysis as described in Eqn.~\ref{eq:eq4} $\sim$ \ref{eq:eq5}. We select Top~$K_2$ OCR tokens according to the attention distribution of Eqn.~\ref{eq:eq5}. 
M4C utilizes the sampled frame features and OCR token features for the QA prediction.

\textbf{T5-ViteVQA} \cite{zhao2022towards} is a TextVideoQA model. To obtain its spatio-temporal grounding results, we refer to the same method as above for post-hoc attention analysis. We select Top~$K_1$ $\times$ $K_2$ OCR tokens from all OCR tokens of the video according to the attention distribution. Similarly, the result of post-hoc attention does not affect QA prediction; T5-ViteVQA uses all frame features and OCR token features for QA prediction.

\textbf{MIST} \cite{gao2023mist} is a VideoQA model. Since general VideoQA models struggle to understand scene text in videos, we extend MIST to the Grounded TextVideoQA task by replacing its original answer decoder with the answer decoder proposed in this paper. For spatio-temporal grounding, MIST first divides the video into segments of equal length, then introduces a cascaded segment and region selection module to adaptively select a question-related video segment, and further selects Top $K_1$ frames and Top $K_2$ OCR tokens. In implementation, we use the grounded frames and OCR tokens to perform QA.

\textbf{TranSTR} \cite{li2023discovering} is a VideoQA model. We also apply our answer decoder to TranSTR to extend it to the Grounded TextVideoQA. For spatio-temporal grounding, TranSTR designs a differentiable selection module that adaptively collects Top $K_1$ critical frames and Top $K_2$ OCR tokens in each frame as grounded results, and finally uses OCR-enhanced frame features for QA prediction.

{
\textbf{Qwen2-VL~\cite{wang2024qwen2}} is a cutting-edge open-source MLLM that supports video inputs, excelling in VQA and visual grounding tasks. By integrating temporal and spatial information, it delivers strong performance in complex spatio-temporal reasoning and grounding scenarios. To explore the performance of Qwen2-VL on the Grounded TextVideoQA task, we design a dedicated Chain-of-Thought prompt (CoT) (Fig. \ref{fig:r1}) to explore its QA and grounding capabilities. The CoT prompt guides the model in identifying the most relevant video frame and scene text bounding box for answering the question.
}

{
\textbf{GPT-4o-mini~\cite{gpt4omini2024}} is a closed-source, lightweight variant of the GPT-4 family, designed for efficiency while maintaining robust multimodal reasoning capabilities. Although it performs well in VQA, its visual grounding ability is constrained by weaker spatio-temporal reasoning. As with Qwen2-VL, using the designed CoT prompt in Fig. \ref{fig:r1}, we also evaluate GPT-4o-mini's performance on the Grounded TextVideoQA task.
}

\begin{table}[t]
\caption{Ablation study of temporal-spatial grounding.}
\label{tab:tab4}
\setlength{\tabcolsep}{.4em}
\fontsize{10}{13}\selectfont
\resizebox{\columnwidth}{!}{
\begin{threeparttable}
\begin{tabular}{l|cccccc}
\Xhline{1pt}
\makecell[c]{Variant} & \makecell[c]{IoU@0.5} & \makecell{IoU@0.3} & \makecell[c]{Acc} & \makecell[c]{ANLS}  & \makecell[c]{GQA@0.5} & \makecell{GQA@0.3}  \\
\Xhline{1pt}
\textbf{\model}  & \textbf{28.14} & \textbf{41.62}  & \textbf{10.20} & \textbf{0.1472} & \textbf{2.96} & \textbf{3.62}   \\
\hline
w/o TSG &  - & - & 8.64 & 0.1380 & - & - \\ 
w/o SG &  3.35 & 5.77  & 8.94 & 0.1351 & 0.54 &  0.63  \\ 
w/o TG  & 14.85 & 25.55 & 9.57 & 0.1430 & 1.96 & 3.05 \\ 
\Xhline{1pt}
\end{tabular}
\end{threeparttable}}
\end{table}

\begin{table}[t]
\caption{Ablation study of the input of the answer decoder.}
\label{tab:tab6}
\setlength{\tabcolsep}{.4em}
\fontsize{10}{13}\selectfont
\resizebox{\columnwidth}{!}{
\begin{threeparttable}
\begin{tabular}{l|cccccc}
\Xhline{1pt}
\makecell[c]{Variant} & \makecell[c]{IoU@0.5} & \makecell{IoU@0.3} & \makecell[c]{Acc} & \makecell[c]{ANLS} & \makecell[c]{GQA@0.5} & \makecell{GQA@0.3}  \\
\Xhline{1pt}
\textbf{\model}  & \textbf{28.14} & \textbf{41.62}  & \textbf{10.20} & \textbf{0.1472} & \textbf{2.96} & \textbf{3.62}   \\
\hline
w/o O  & 16.98  & 25.69 &  5.44 &  0.0803 & 1.27   & 0.63  \\
w/o F & 25.68 & 37.08  & 9.72 & 0.1391 & 2.05 & 2.87 \\ 
w/o F \& Q & 23.93 & 34.93 & 8.57 & 0.1179 &  1.87 & 2.69 \\ 
\Xhline{1pt}
\end{tabular}
\end{threeparttable}}
\end{table}

%------------------------------------------------------
\begin{figure}[t]
\centering
\includegraphics[width=\columnwidth]{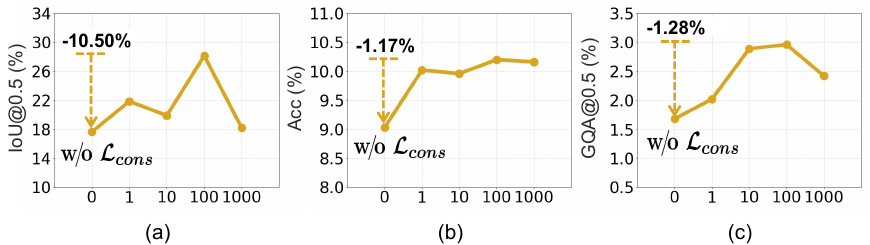}
\caption{Study of contrastive loss $\mathcal{L}_{cons}$ with $\lambda$.}
\label{fig:fig6}
\end{figure}

The main results in Tab.~\ref{tab:tab3} show that our model \model ~consistently outperforms all baselines under different evaluation metrics and settings. Such strength is unique to \model compared with other baseline methods. More specifically, \model ~wins more under the Top-5 setting versus the Top-1, especially for grounding. Examining the performance of all other methods under the Top-1 setting, we find that accurately localizing the precise scene-text region corresponding to a specific answer is highly challenging. This also explains the relatively marginal performance gap among models.

{
In Tab.~\ref{tab:r1_mllm}, we evaluate the performance of MLLMs on \dataset~through three setups: direct QA (“w/ Standard”) in Fig.~\ref{fig:r1} (a), QA with Chain-of-Thought (CoT) prompt (“w/ CoT”) in Fig.~\ref{fig:r1} (b), and scene-text grounded QA with Chain-of-Thought prompt (“w/ Grounded CoT”) in Fig.~\ref{fig:r1} (c). Our findings are as follows: 
(1) \emph{MLLMs face severe challenges in scene-text spatio-temporal grounding despite their good QA performance.} The results show that under the “Grounded CoT” strategy, GPT-4o-mini~\cite{gpt4omini2024} and Qwen2-VL~\cite{wang2024qwen2} achieve low IoU@0.5 scores of 0.19\% and 2.14\%, respectively, failing behind T2S-QA. (2) \emph{CoT prompting improves QA accuracy.} Switching from “w/ Standard” to “w/ CoT”, GPT-4o-mini~\cite{gpt4omini2024} shows a 3.80\% increase in Acc, demonstrating CoT’s effectiveness in enhancing reasoning. To further explore the CoT prompt, we adopt a representative QA pair as the one-shot CoT prompt in Fig.~\ref{fig:r1} (c). The results demonstrate a modest improvement in both textual answer accuracy and visual grounding performance. (3) \emph{The performance of explicit grounded reasoning and implicit grounded reasoning is inconsistent.} When transitioning from “w/ Standard” to “w/ Grounded CoT”, Qwen2-VL~\cite{wang2024qwen2} suffers a 15.27\% decrease in Acc, indicating the inability to effectively integrate grounding in the QA process. We further identify key QA failure cases, including reliance on irrelevant scene text or hallucinated visual content and difficulty recognizing occluded or partially visible text, underscoring the challenges and future directions in scene text-aware video QA.}

%------------------------------------------------------
\begin{table}[t]
\centering
\caption{{ Ablation study of different scene text features. \emph{Linguistic} features include FastText feature and PHOC feature. \emph{Spatial} features include bounding box features, \emph{Temporal} features include temporal id features and track id features. }}
% \vspace{-0.1in}
\label{tab:r1_feat}
\setlength{\tabcolsep}{.4em}
\fontsize{10}{13}\selectfont
\resizebox{\columnwidth}{!}{
\begin{tabular}{c|c|c|ccccc}
\Xhline{1pt}
Linguistic & Spatial & Temporal  & \makecell[c]{IoU@0.5} & \makecell[c]{Acc} & \makecell[c]{ANLS}  & \makecell[c]{GQA@0.5}  \\
\Xhline{1pt}
\checkmark & - & - &   24.64 & 10.30 & 0.1414  & 3.05  \\
\checkmark & \checkmark & - & 28.31 & 10.39 & 0.1471  & 2.60 \\
\checkmark & \checkmark & \checkmark &  \textbf{28.60} & \textbf{10.45} & \textbf{0.1520} & \textbf{3.50} \\
\bottomrule
\end{tabular}}
\vspace{-0.15in}
\end{table}

%------------------------------------------------------
\begin{table}[t]
\caption{Performance with different numbers of video frames. Box Cov. (\%) is the average coverage percentage of annotated boxes w.r.t the sampled video frames. } 
\label{tab:suptab2}
\setlength{\tabcolsep}{.3em}
\fontsize{10}{13}\selectfont
\resizebox{\columnwidth}{!}{
\begin{threeparttable}
\begin{tabular}{c|c|cccccc}
\Xhline{1pt}
\makecell[c]{$T$} & \makecell[c]{Box Cov.} & \makecell[c]{IoU@0.5} & \makecell{IoU@0.3} & \makecell[c]{Acc} & \makecell[c]{ANLS} & \makecell[c]{GQA@0.5} & \makecell{GQA@0.3}  \\
\Xhline{1pt}
32 & 68.75 & 18.79 & 27.95 & 9.55  &  0.1418 &  1.87  & 2.68  \\ 
48 & 82.19 & 18.84 & 27.55 & 9.66 & 0.1413  &  1.96 & 2.87 \\ 
\textbf{64} & \textbf{92.25} &\textbf{28.14} & \textbf{41.62}  & \textbf{10.20} & \textbf{0.1472} & \textbf{2.96} & \textbf{3.62} \\
80 & 96.62 & 18.68 & 27.09 & 9.93 & 0.1413 & 2.51  & 2.92 \\ 
\Xhline{1pt}
\end{tabular}
\end{threeparttable}}
\end{table}

%------------------------------------------------------
\begin{table}[t]
\centering
\caption{{Performance with different numbers of transformer layers.}}
\label{tab:r1_trans}
\setlength{\tabcolsep}{.3em}
\fontsize{12}{14}\selectfont
\resizebox{\columnwidth}{!}{
\begin{tabular}{l|cccccc}
\Xhline{1pt}
\makecell[c]{Variant} & \makecell[c]{IoU@0.5} & \makecell{IoU@0.3} & \makecell[c]{Acc} & \makecell[c]{ANLS}  & \makecell[c]{GQA@0.5} & \makecell{GQA@0.3}  \\
\Xhline{1pt}
w/o Trans. Layer & 16.36 & 25.72 & 9.18 &  0.1272  & 2.14 & 2.42  \\
w/ 1-layer Trans. & 25.08 & 38.18 & 8.70 & 0.1262 & 2.87 & 4.23   \\
\textbf{w/ 2-layer Trans.} & \textbf{28.60} & \textbf{41.94} & \textbf{10.45} & \textbf{0.1520} & \textbf{3.50}  & \textbf{ 4.23 } \\
w/ 3-layer Trans. & 27.00 & 39.53 & 8.97 & 0.1294 & 3.17 & 3.89   \\
\bottomrule
\end{tabular}}
\vspace{-0.1in}
\end{table}

\subsection{{Ablation Study (RQ2)}}
To understand the effects of different components in \model, we dissect the model performance on \dataset~test set and report results with the Top 5$\times$5 setting.

\noindent\textbf{Effectiveness of Temporal-to-Spatial Grounding.}
In Tab. \ref{tab:tab4}, we respectively study the effectiveness of \model~by removing temporal and spatial grounding (“w/o TSG”), only removing temporal grounding (“w/o TG”), and only removing spatial grounding (“w/o SG”). Specifically, “w/o TG” variant uses the grounded OCR tokens in all video frames as positive OCR tokens and the others as negative OCR tokens;
“w/o SG” variant uses all OCR tokens of the positive video frame as the positive OCR token, and all OCR tokens of the negative video frame as the negative OCR token;
“w/o TSG” variant represents reserving all video frames and OCR tokens for answering, ${\textit{i.e.}}$, without answer grounding.

Tab.~\ref{tab:tab4} shows that both spatial and temporal grounding are key for better QA performance.
We note that the model variant “w/o SG” performs the worst in both answer grounding and grounded QA. This suggests that only learning the differences between video frames is insufficient to capture subtle local changes in scene text within a frame, making it challenging to achieve accurate answer grounding. Moreover, compared with T2S-QA, “w/o TG” also shows a significant decrease of 13.29\% on IoU@0.5 and 1\% decrease in GQA@0.5. This indicates that directly performing one-stage grounding of scene text across the entire video yields suboptimal results.

\noindent\textbf{Study Inputs of Answer Decoder.}
Tab.~\ref{tab:tab6} shows that when removing OCR tokens (``w/o O'') both \model's grounding and QA results degenerate significantly by 11.16\% on IoU@0.5 and 4.76\% in Acc respectively. This demonstrates the significance of OCR token grounding for answer prediction. Additionally, we show that both the frame appearance F feature and question Q help the performance.

\noindent\textbf{Effectiveness of Contrastive Loss.}
Fig.~\ref{fig:fig6} shows that \model's performances degenerate remarkably without the contrastive loss, especially the grounding result degrades by over 10\%. We also investigate different values for the coefficients $\lambda$. The results suggest that $\lambda$ = 100 brings the best performance.

{
\noindent\textbf{Study of Scene Text Features.} In Tab.~\ref{tab:r1_feat}, the results show that linguistic features obtained by character-level encoding are crucial based on the scene text recognition results. Capturing spatial and temporal features can complement the spatiotemporal characteristics of scene text in videos. Their combination produces the best performance.
}

\noindent\textbf{Different Numbers of Video Frames.}
We conduct experiments with four different numbers of sampled video frames, \ie, $T$ = 32, 48, 64, and 80. 
Tab. \ref{tab:suptab2} shows that uniformly sampling 80 frames can cover over 96\% of the annotated scene-text regions. However, the best results are achieved at frame number of 64. We speculate that learning with too many frames make it harder to identify the very limited positive frames that carry the needed scene texts.

{
\noindent\textbf{Different Numbers of Transformer Layers.} In Tab.~\ref{tab:r1_trans}, we conduct experiments by varying the number of Transformer layers for the architecture introduced at Sec.~\ref{sec:fea_Rep}. It shows that when the Transformer module is
removed (“w/o Trans. Layer”), the performance on IoU@0.5 drops by 12.24\%, and the performance
on GQA@0.5 drops by 1.03\%. The results show that the optimal
configuration is with two layers.
}

%------------------------------------------------------
\begin{table}[t]
\caption{Study of scene-text grounding and recognition in Grounded TextVideoQA.} % 
%\vspace{-0.08in}
\label{tab:suptab4}
\setlength{\tabcolsep}{.2em}
\fontsize{12}{14}\selectfont
\resizebox{\columnwidth}{!}{
\begin{threeparttable}
\begin{tabular}{l|cccccc}
\Xhline{1pt}
\makecell[c]{Method} & \makecell[c]{IoU@0.5} & \makecell{IoU@0.3} & \makecell[c]{Acc} & \makecell[c]{ANLS}  & \makecell[c]{GQA@0.5} & \makecell{GQA@0.3}  \\
\Xhline{1pt}
Human  & 77.39  & 80.90 & 64.32 & 0.8012 &  51.76 & 54.27  \\
Upper Bound & 56.52 & 76.07 & 72.62 & 0.6967 & 41.53 & 56.15   \\
\hline
\model~(CLIPOCR) & {28.60} & {41.94} & {10.45} & {0.1520} & {3.50}  & { 4.23 } \\
{w/ CLIPOCR GT Box}  & 96.70 &  96.81 & 11.96 & 0.1875 & 11.24 & 11.56  \\ 
\Xhline{1pt}
\end{tabular}
\end{threeparttable}}
\end{table}

%------------------------------------------------------
\begin{table}[t]
\caption{{Performance of \model~under different scene-text detection and recognition systems. }}
\label{tab:reb-tab3}
\setlength{\tabcolsep}{.2em}
\fontsize{12}{14}\selectfont
\resizebox{\columnwidth}{!}{
\begin{threeparttable}
\begin{tabular}{l|c|ccccc}
\Xhline{1pt}
OCR Detection & OCR Recognition  &  \makecell[c]{IoU@0.5}  & \makecell[c]{Acc} & \makecell[c]{ANLS}  & \makecell[c]{GQA@0.5} \\
\Xhline{1pt}
GoMatching~\cite{he2024gomatching} & ABINet \cite{fang2021read} & {25.26}  & {9.89} & {0.1436} & {2.15}   \\
TransVTSpotter \cite{wu2021bilingual} & ABINet \cite{fang2021read}  & {28.14} & {10.20} & {0.1472} & {2.69}  \\
TransVTSpotter~\cite{wu2021bilingual} & {CLIPOCR}~\cite{wang2023symmetrical} & \textbf{28.60}  & \textbf{10.45} & \textbf{0.1520} & \textbf{3.50}   \\
\Xhline{1pt}
\end{tabular}
\end{threeparttable}}
\vspace{-0.1in}
\end{table}

%------------------------------------------------------
%\vspace{-0.1in}
\subsection{{In-Depth Analysis (RQ3)}} 
Our previous experiments show that models obtain low QA accuracy even with relatively higher grounding behavior. We speculate that the reasons are two folds: \emph{ineffective answer evaluation} and \emph{poor scene-text detection and recognition}. The former indicates that the predicted textual answer does not exactly match the ground-truth one (\eg~30 \vs~30 M.P.H in Fig.~\ref{fig:fig1}). The latter hints that the OCR system fails to convert the grounded scene-text into textual answers (\eg~M \vs~M.P.H).

\noindent\textbf{Human Performance and Upper Bound.}
To discern the error from ineffective answer evaluation, we sample {around 20\% test data which contains 200 questions and 163 videos} and conduct a human study by asking volunteers to answer the questions and ground the related scene text areas. Table \ref{tab:suptab4} shows that even human (we assume that humans are with perfect scene-text recognition) can only achieve an accuracy of 64.32\%. A further investigation of the the wrong answers suggests that an absolute 25\% out of 36\% of they are acceptable and should be regarded as correct. For example, ``35131.86'' \vs~``35,131.86''. The results suggest that stringent string matching is ineffective in assessing TextVQA models. Yet, humans' significantly higher grounding performance indicates that grounding annotation is \emph{less subjective}, highlighting the importance of directly evaluating scene-text regions. The gap between our model and human performance underscores the need for further research into trustworthy TextVideoQA.

We report the ``Upper Bound” result with the TransVTSpotter~\cite{wu2021bilingual} detection and CLIPOCR~\cite{wang2023symmetrical} recognition systems, obtained via brute-force search over all detected text regions and recognized outputs. The result in Table \ref{tab:suptab4} suggests that even a brute-force search can only achieve a grounding accuracy of 56.52\%. Note that \model~achieves 28.60\%. This reflects the effectiveness of our grounding approach, though the accuracy is not so high. Additionally, we find that the ``Upper Bound” is inferior to human in grounding but exceeds human in QA accuracy.
This reveals the inadequacy of the existing OCR system in scene-text detection and the challenge of effectively evaluating textual answers because of human subjectivity.

{\noindent\textbf{Impact of Scene Text Detection and Recognition.} As shown in Table~\ref{tab:reb-tab3}, we evaluate various scene text detection and recognition systems and analyze their impact on the Grounded TextVideoQA task from three perspectives. 1) To assess the impact of OCR detection systems, we test the variant where the ground-truth scene text for each QA pair is directly provided to the answer decoder.
{Compared to the human-annotated answer grounding result (“w/ CLIPOCR GT Box”), the performance of directly iterating over all OCR detection results (“Upper Bound”) drops by 40.18\% on IoU@0.5. This drop illustrates the difficulty of scene text detection in dynamic video environments. }
2) We analyze the impact of OCR recognition systems. The GQA@0.5 score of ``T2S-QA (CLIPOCR)" is 25.1\% which is lower than its IoU@0.5 score, suggesting incorrect textual answers even when the scene text is located accurately. The comparison between ``T2S-QA (CLIPOCR)" and ``w/ CLIPOCR GT Box" shows only marginal improvement (1.51\% in Acc) despite perfect grounding. These results emphasize the limitations of current video-based scene text recognition systems.
3) We evaluate various combinations of OCR detection and recognition systems. Using the same recognition system, TransVTSpotter achieves superior grounding due to its robust detection capabilities. Similarly, employing CLIPOCR as the recognition method significantly improves grounding accuracy when using a fixed detection system. These results demonstrate that high-quality detection-recognition combinations enhance grounding performance and optimize QA accuracy.}

%------------------------------------------------------
\begin{figure*}
\centering
\includegraphics[width=\textwidth]{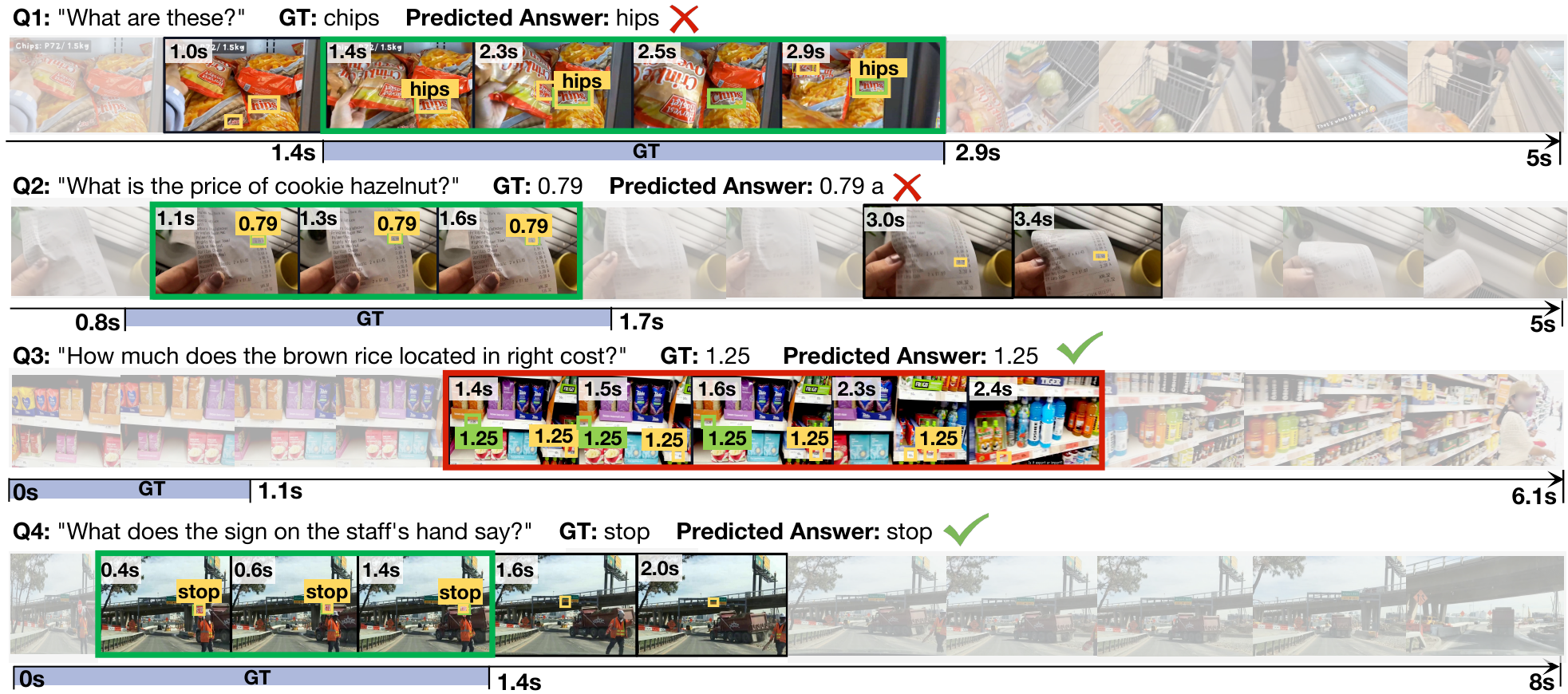}
% \vspace{-0.2in}
\caption{Case-Study of \model~on \dataset~test set. The green and yellow boxes in video frames denote the ground-truth boxes and the model-grounded OCR tokens. The blue bar denotes the annotated temporal segment.
}
\label{fig:fig5}
\vspace{-0.1in}
\end{figure*}

\noindent\textbf{Qualitative analysis.} As shown in Fig. \ref{fig:fig5}, in \textbf{Q1}, \model~successfully identifies four frames at 1.4s, 2.3s, 2.5s, and 2.9s, and then pinpoints the answer region in frames 1.4s, 2.3s, and 2.9s. However, due to insufficient light and partial occlusion, an error occurred in scene text recognition where ``chips'' is mistakenly recognized as ``hips'' leading to an incorrect model answer.  In \textbf{Q2}, \model~correctly locates the key video frames and scene text ``0.79'' for answering the question. However, due to a one-word difference between the predicted answer and the ground truth answer, with an extra word ``a'', the predicted answer is considered incorrect. In \textbf{Q1} and \textbf{Q2}, the grounded visual answers served as valuable evidence in diagnosing the issues with the model. 
In \textbf{Q3}, although the model answers the question correctly, it does not correctly locate the price of brown rice on the right, but locates other products with the same price. Therefore, the answer may be based on the price of other products ``1.25'' on the incorrectly identified video frames of 1.4s, 1.5s, 1.6s, 2.3s, and 2.4s.
In \textbf{Q4}, \model~correctly locates and recognizes the scene text ``stop'' for answering the question.
In conclusion, \model~provides visual evidence for answer prediction, enhancing the transparency of the decision-making process. It assists us in diagnosing the underlying causes of failure prediction more effectively, thereby enabling better model optimization.

\vspace{-0.1in}
\section{Conclusion}
\label{sec:conclu}
We explore a novel task called Grounded TextVideoQA. The task is crucial for developing trustworthy VideoQA systems involving scene texts. The task also allows for better evaluation of models by enabling direct assessment of grounded scene text regions instead of just textual answers. To solve the task, we propose \model, a visually grounded TextVideoQA model featuring a disentangled temporal-to-spatial contrastive grounding mechanism for scene text grounding and grounded TextVQA. Additionally, we construct the \dataset~benchmark to facilitate evaluation, which enables direct evaluation of spatio-temporal scene-text grounding. Comparative results with existing models and ablated variants demonstrate the superiority and effectiveness of our approach in both QA and answer grounding. Moreover, we conduct an in-depth analysis of the model's specific behavior concerning grounding and QA accuracy. The results indicate that the primary reasons for poor TextVideoQA performance are the sub-optimal OCR system and ineffective evaluation metrics. We hope our dataset, baselines, and analyses will contribute to the advancements in faithfully answering scene-text questions in videos.

%%%%%%%%% REFERENCES
{
\bibliographystyle{ieee_fullname}
\bibliography{ref}
}

\end{document}